\documentclass[letterpaper, 10 pt, conference]{ieeeconf}  % Comment this line out if you need a4paper

\IEEEoverridecommandlockouts                              % This command is only needed if 
\usepackage{url}
\title{\LARGE \bf
Neural Implicit Representation for Highly Dynamic LiDAR Mapping and Odometry
}

\author{Qi Zhang$^{1,*}$, He Wang $^{2,*}$, Ru Li$^{2}$, and Wenbin Li$^{1}$% <-this % stops a space
\thanks{$^{*}$These authors contributed equally and are considered as co-first authors}% <-this % stops a space
\thanks{$^{1}$Department of Computer Science,
        University of Bath, UK, \{\protect\url{qz727, w.li}\}\protect\url{@bath.ac.uk}}%
\thanks{$^{2}$School of Computer and Information Technology, ShanXi University, ShanXi, China, {\protect\url{wh_sxu@foxmail.com}, \protect\url{liru@sxu.edu.cn}}}%
}

\usepackage{graphicx}
\usepackage{lipsum}
\usepackage{tabularx}
\usepackage{array}
\usepackage{amsmath}
\usepackage{amssymb}
\usepackage{float}
\usepackage{balance}
\usepackage{color}
\usepackage{svg}

\usepackage{subcaption}  %插入多图时用子图显示的宏包

\begin{document}

\maketitle
\thispagestyle{empty}
\pagestyle{empty}

%%%%%%%%%%%%%%%%%%%%%%%%%%%%%%%%%%%%%%%%%%%%%%%%%%%%%%%%%%%%%%%%%%%%%%%%%%%%%%%%
\begin{abstract}

Recent advancements in Simultaneous Localization and Mapping (SLAM) have increasingly highlighted the robustness of LiDAR-based techniques. At the same time, Neural Radiance Fields (NeRF) have introduced new possibilities for 3D scene reconstruction, exemplified by SLAM systems. Among these, NeRF-LOAM has shown notable performance in NeRF-based SLAM applications. However, despite its strengths, these systems often encounter difficulties in dynamic outdoor environments due to their inherent static assumptions.
To address these limitations, this paper proposes a novel method designed to improve reconstruction in highly dynamic outdoor scenes. Based on NeRF-LOAM, the proposed approach consists of two primary components. First, we separate the scene into static background and dynamic foreground. By identifying and excluding dynamic elements from the mapping process, this segmentation enables the creation of a dense 3D map that accurately represents the static background only.
The second component extends the octree structure to support multi-resolution representation. This extension not only enhances reconstruction quality but also aids in the removal of dynamic objects identified by the first module. Additionally, Fourier feature encoding is applied to the sampled points, capturing high-frequency information and leading to more complete reconstruction results.
Evaluations on various datasets demonstrate that our method achieves more competitive results compared to current state-of-the-art approaches.

\end{abstract}

%%%%%%%%%%%%%%%%%%%%%%%%%%%%%%%%%%%%%%%%%%%%%%%%%%%%%%%%%%%%%%%%%%%%%%%%%%%%%%%%
\section{INTRODUCTION}
% 1. 激光雷达+双目slam
% 2. NeRF在slam的应用 有些纯雷达，有些雷达+rgb，但是outdoor上其实很多动态物体，比如车和人，很多方法使用的数据集都是静态的街道，对于动态物体多的街道的重建很差。
% 3. 我们的方法基于NeRF loam，在它的基础上认为动态物体是outlier，再填补空洞

Recently, Neural Radiance Fields (NeRF) \cite{mildenhall2021nerf} have emerged as a powerful tool for 3D scene reconstruction and novel view synthesis, largely due to their ability to generate highly detailed and realistic representations of complex scenes from sparse observations. NeRF's strength lies in its capacity to model scenes as continuous volumetric fields, enabling high-quality novel view synthesis and accurate 3D reconstructions from multiple input images. This capability makes NeRF particularly promising for Simultaneous Localization and Mapping (SLAM) applications, both indoors and outdoors. By capturing intricate scene details and providing dense volumetric representations, NeRF enhances SLAM systems, improving spatial understanding and object recognition.
In indoor environments, several NeRF-based SLAM systems have been developed, including iMap \cite{sucar2021imap}, NICE-SLAM \cite{zhu2022nice}, NeRF-SLAM \cite{rosinol2023nerf}, Co-SLAM \cite{wang2023co}, GO-SLAM \cite{zhang2023go}, and NGEL-SLAM \cite{mao2023ngel}. These systems leverage camera sensors to deliver precise reconstructions. For outdoor scenarios, NeRF-based SLAM has been extended to incorporate LiDAR sensors, as demonstrated in NeRF-LOAM \cite{deng2023nerf}, LONER \cite{isaacson2023loner}, and PinSLAM \cite{pan2024pin}. Additionally, some systems combine visual and LiDAR data, such as CLONeR \cite{carlson2023cloner} and SiLVR \cite{tao2024silvr}.
However, most NeRF-based SLAM systems operate under the assumption that the environment is static or only minimally dynamic. This assumption presents significant challenges when applying these methods to the real world with highly dynamic objects in outdoor scenarios, where the scene reconstruction becomes considerably inaccurate.

\begin{figure}[tp]
    \centering
    \vspace{8pt}
    \begin{subfigure}[b]{0.471\textwidth}
        \centering
        \includegraphics[width=\textwidth]{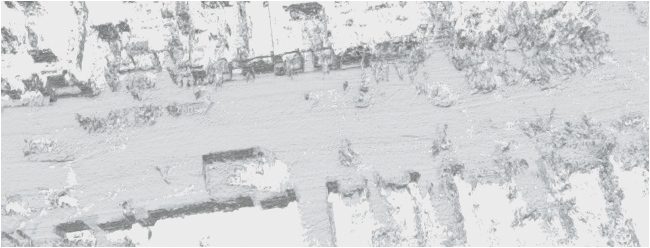}
        \caption{Ours in KITTI MOT 19}
        \label{our19}
    \end{subfigure}
    \begin{subfigure}[b]{0.471\textwidth}
        \centering
        \includegraphics[width=\textwidth]{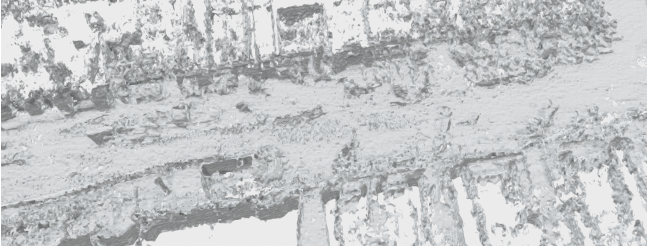}
        \caption{NeRF-LOAM \cite{deng2023nerf} in KITTI MOT 19}
        \label{ori19}
    \end{subfigure}
    \caption{The 3D reconstruction of KITTI MOT 19 using our proposed method and NeRF-LOAM \cite{deng2023nerf}. 
    % (b) The 3D reconstruction using NeRF-LOAM \cite{deng2023nerf} is presented for comparison. Our method demonstrates a significant ability to remove moving objects from the scenarios, resulting in a much cleaner and more accurate map. 
    }
    \label{Fig:introduction}
\end{figure}

In this paper, we aim to construct a dense 3D map of outdoor scenes in highly dynamic scenarios. To achieve this, we enhance NeRF-LOAM \cite{deng2023nerf} by integrating an additional thread that detects the moving objects, generating their 3D bounding boxes. This allows us to categorize the LiDAR points into background and foreground segments, under the assumption that the background remains static while the foreground is dynamic. Building upon the foundation of NeRF-LOAM \cite{deng2023nerf}, we focus exclusively on calculating the Signed Distance Function (SDF) values for the background, ensuring that dynamic elements are accurately separated from the static environment.

Moreover, we propose a hybrid feature representation method that significantly benefits the generation of highly dynamic objects. We extended the octree structure in NeRF-LOAM \cite{deng2023nerf} to support multiple resolutions, similar to the approach used in NGLOD \cite{takikawa2021neural}. By considering multiple resolutions, our method can better capture the fine details and motion of dynamic objects across different scales. Inspired by Co-SLAM \cite{wang2023co}, we recognized that although parametric encoding alone can improve reconstruction results, it has limitations in filling holes and ensuring smooth transitions.
To overcome these challenges, we combine multiple learnable features with Fourier feature positional encoding \cite{tancik2020fourier}. 

Our experimental results demonstrate that our methods are particularly effective for highly dynamic scenes, as it not only remove the dynamic objects and fills holes, but also enhances smoothness, which is crucial for maintaining the integrity and continuity of fast-moving objects. The key contributions are summarized as follows:

\begin{itemize}
    \item  We enhance NeRF-LOAM \cite{deng2023nerf} by integrating a method to segment scenarios into the dynamic foreground and static background, removing dynamic LiDAR points and rebuilding ground points to facilitate accurate 3D mapping in highly dynamic outdoor environments.

    \item We extended the single-layer learnable features in the octree of NeRF-LOAM \cite{deng2023nerf} to multiple layers and applying Fourier feature encoding to the sampled points, allowing us to achieve better reconstruction results.

    \item To improve the accuracy of SDF values in highly dynamic scenarios, we incorporate additional loss functions into the optimization process.

\end{itemize}

\section{RELATED WORKS}

% 首先提到了lidar based slam  然后是 rgb based slam  最后是多传感器fusion的slam
Simultaneous Localization and Mapping (SLAM) is a foundational technology in robotics and autonomous driving, crucial for enabling machines to navigate and understand their surroundings. SLAM systems can be categorized based on the sensors they utilize. LiDAR-based SLAM methods, such as Lo-net \cite{li2019net}, Deeppco \cite{wang2019deeppco}, and Pwclo-net \cite{wang2021pwclo}, have gained prominence due to their robustness and illumination invariance. Relying on RGB cameras, SLAM systems can capture color and texture information, including ORB-SLAM2 \cite{mur2015orb}, ORB-SLAM3 \cite{mur2017orb}, and DynaSLAM \cite{bescos2018dynaslam, bescos2021dynaslam}. Combining LiDAR and RGB data, like Gaussian-LIC \cite{lang2024gaussian}, leverages the strengths of both modalities.

% 传统的隐式表示 显式表示 最后引入可微渲染函数
In parallel with advancements in SLAM, Neural Implicit Representations (NeRF) \cite{mildenhall2021nerf} have significantly influenced 3D scene reconstruction techniques. Early approaches mapped point coordinates to signed distance functions (SDFs) \cite{park2019deepsdf} and occupancy fields \cite{mescheder2019occupancy}, but these methods required access to ground truth 3D geometry, which limited their applicability. VDBFusion \cite{vizzo2022vdbfusion} stores TSDF values in sparse voxels, resulting in a highly accurate but incomplete reconstructed map, Puma \cite{vizzo2021poisson} represents the map as a triangle mesh through Poisson reconstruction, enabling it to capture more detailed maps compared to common mapping methods. More recent work introduced differentiable rendering functions to represent scenes using only 2D images \cite{niemeyer2020differentiable, sitzmann2019scene}. While these approaches improved accessibility to 3D scene reconstruction, they often resulted in overly smooth representations for complex shapes \cite{zhong2023shine}.

% nerf based slam
In the realm of NeRF-based SLAM, several approaches have been proposed to integrate NeRF with SLAM systems. IMap \cite{sucar2021imap} was an early effort in this direction but struggled with network-induced forgetting and capacity limitations in large-scale environments. NICE-SLAM \cite{zhu2022nice} improved on this by subdividing the scene into uniform grids and using a pre-trained geometry decoder for better reconstruction. ESLAM \cite{johari2023eslam} employed axis-aligned feature planes to manage memory requirements and used an implicit Truncated Signed Distance Field (TSDF) \cite{azinovic2022neural} for improved geometry representation and faster convergence. Vox-Fusion \cite{yang2022vox} introduced an incremental scene representation using octrees, which leveraged embeddings at octree leaf nodes for interpolation.

% 大规模场景下 nerf based slam
For large-scale environments, NeRF-LOAM \cite{deng2023nerf} represents the first attempt to combine neural implicit representations with LiDAR data for odometry and mapping. Despite its advancements, NeRF-LOAM \cite{deng2023nerf} faces challenges with dynamic object removal and hole filling. Pin-SLAM \cite{pan2024pin} uses optimizable neural points to construct an implicit map, but the improvement in reconstruction quality is limited by the fixed resolution of the neural points.

% 为了解决问题我们提出的方法
To address these issues, our approach involves removing points from dynamic regions, performing static filling to close large gaps, and reconstructing the scene using multi-resolution and Fourier feature encoding, thereby enhancing the overall robustness and accuracy of the reconstruction.

\section{METHOD}

\subsection{System Overview}

\begin{figure*}
     \centering
     \includegraphics[width=0.9\textwidth]{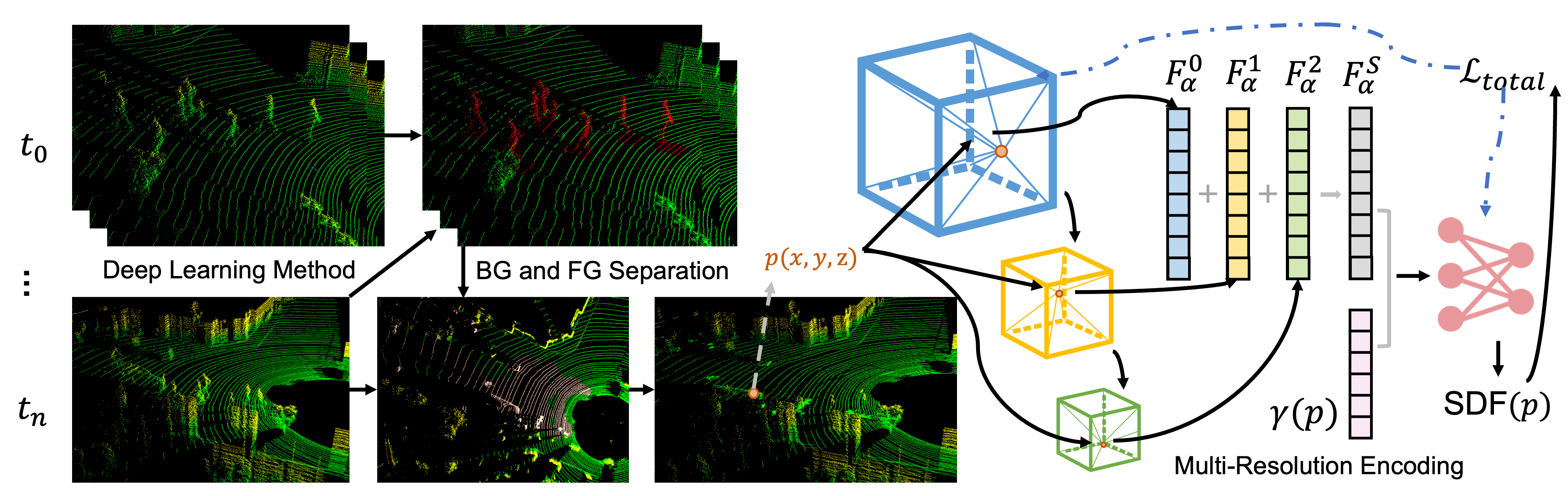}
     
     \caption{The overview of the system. The left part of the image illustrates the process of background and foreground separation.  We remove the dynamic points from the foreground, and generate a foreground mask (pink), resulting in a purely static scene.
     % In each frame $t_i$, where $i \in [0, n]$, we use a deep learning-based method to detect and track moving objects, and mask them in the scene, which are represented as red points. Then, we aggregate all the masks from each frame to construct a comprehensive dynamic foreground, shown as the pink points in the middle-bottom of the left part of the image. Then, we remove these points from the point cloud, resulting in a static scene that excludes all moving objects.
     The right part of the image shows the training process of the neural SDF module. We interpolate the query point at different resolution levels to obtain the corresponding features, and finally combine the Fourier feature positional encoding and feed them into the MLP to predict the SDF value.
     % , $F_{\alpha}^{i} \left \{ i = 0,1,2 \right \}$, each represents the embedding obtained from the interpolation of different nodes, which are then concatenated to form $F_{\alpha}^{s}$, $F_{\alpha}^{s}$ and Fourier encoding of the sampled point $\gamma \left (  q\right )$ are fed into the network to predict the SDF value of point p, denoted as $SDF(p)$. We use $\mathcal{L}_{\text{total}}$ to optimize the network parameters and the embeddings of the vertices.
     }
    \label{fig:overview}
\end{figure*}

The Figure. \ref{fig:overview} is the overview of the proposed system. It can be separated into two sections. The left part of the image shows the background and foreground separation, which is illustrated in Section \ref{bg_and_fg_separation}. In each frame, we detect the moving objects, then combine them with the moving objects from previous frames to form the foreground. For the points in the foreground, if they are not ground points, we remove all of them and generate the ground points within the space. 

The right part of the image illustrates the scene representation and the training process of the SDF values. For the same query point, the interpolated embedding of its corresponding node is obtained from different levels of the octree. This embedding is then concatenated with the coordinates of the sampled point after Fourier encoding, and the combined data is fed into an MLP network to predict the SDF value, as detailed in Section \ref{scene_representation}. Furthermore, for the non-ground points in the foreground, we propose a dynamic region-based SDF loss function to minimize the loss of the SDF value for these points. Section \ref{loss} describes the final loss function.

\subsection{Background and Foreground Separation}
\label{bg_and_fg_separation}
% 1. 使用了images（in another thread）识别动态物体+3Dbox recognition. 
% 2. 将所有的动态物体走过的轨迹作为 dynamic occupation （认为是foreground）
% 3. 如果是dynamic occupation 认为只生成地面。但是生成地面的时候有一个问题，就是在动态物体整个不考虑的时候， 会生成空洞。所以我们在它的寻找地面生成地面z轴高度的一个平均值，并且随机生成地面雷达点，去进行sdf的重建。（这一部分可以考虑是否能创建一个loss值）

To accurately reconstruct dynamic outdoor scenes, we implement a background and foreground separation strategy that effectively distinguishes between static and dynamic points. This process involves detecting the dynamic objects, marking their trajectories of the dynamic occupations as dynamic foreground zones, and ensuring consistent ground surface reconstruction.

\textbf{Moving Object 3D Box Recognition:} 

\begin{figure}[tp]
     \centering

     \includegraphics[width=0.299\textwidth]{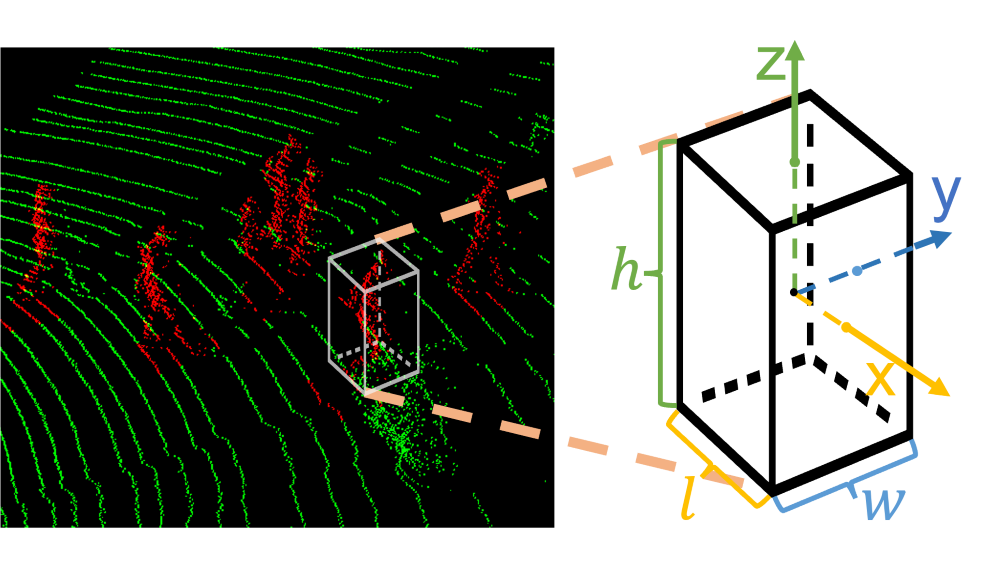}
     \caption{Definition of the 3D box $B$.} 
    \label{fig:3DboxAxis}
\end{figure}

In a separate thread, we use an existed deep-learning-based method to identify moving objects $\left\{O_N, \ldots, O_M\right\} = \left\{O_i\right\}^M_{i = N}$ within the scene. These objects are tracked and enclosed within 3D bounding boxes, defined as $\left\{B_N, \ldots, B_M\right\} = \left\{B_i\right\}^M_{i = N}$. As shown in Figure. \ref{fig:3DboxAxis}, each bounding box $B_i = \{ \boldsymbol{P}_{ci}, h_i, w_i, l_i\}$, where $\boldsymbol{P}_{ci} = \left \{x_{ci}, y_{ci},z_{ci} \right\}^T$ represent the coordinates of the center of the 3D box $B_i$, and $h$ (here the subscript $c$ denotes the center), $w$ and $l$ denote its height, width, and, length respectively.

\textbf{Dynamic Foreground Marking:}

In the outdoor environment, we assume that if moving objects are detected, the space within their corresponding 3D bounding boxes is defined as foreground. During the 3D reconstruction, no structures should be built within the foreground area, except for the ground floor. This separation allows us to focus on reconstructing the static background accurately while treating the moving objects as outliers.

To achieve this, we maintain a global dynamic list  $L$, where each element in the list corresponds to the dynamic mask $D_i$ of a moving object $O_i$. The list is defined as $L = \left\{D_i\right\}^M_{i = N}$, representing the masks for all detected moving objects. The $D_i = \left\{\boldsymbol{p}_{li}, \boldsymbol{p}_{ri}\right\}$ is represented as:
\begin{equation}
    \begin{aligned}
    \boldsymbol{p}_{li} = \left( (\boldsymbol{R} \boldsymbol{P}_{ci})_x + T_x - \frac{w_i}{2},\ (\boldsymbol{R} \boldsymbol{P}_{ci})_y + T_y - \frac{l_i}{2} \right) \\
    \boldsymbol{p}_{ri} = \left( (\boldsymbol{R} \boldsymbol{P}_{ci})_x + T_x + \frac{w_i}{2},\ (\boldsymbol{R} \boldsymbol{P}_{ci})_y + T_y + \frac{l_i}{2} \right)\\
    \end{aligned}
    \label{pliandpri}
\end{equation}
where $(\boldsymbol{R|T})$ is the camera pose with rotation matrix $\boldsymbol{R}$ and translation vector $\boldsymbol{T}$. 

To update the list $L$ , whenever a new frame is input, we also remove elements by checking the $y_{ci}$. If $y_{ci} < T_y$, the corresponding element is removed from the list. 

Thus, when a new frame is input, if the LiDAR points are not ground points and fall within any mask $D_i$ in the list $L = \left\{D_i\right\}^M_{i = N}$, these points are considered as part of the dynamic foreground and not be included in the static background reconstruction. (To check if the LiDAR points are ground points, we use the same method in \cite{deng2023nerf}.)

\textbf{Ground Surface Generation in Foreground:} 

% When dealing with dynamic foreground zones, our method assumes that only the ground surface needs to be generated. However, ignoring dynamic objects entirely could lead to gaps in the ground's reconstruction. To address this, we estimate an average height for the ground surface in the z-axis within these dynamic regions. We then randomly generate LiDAR points at this estimated height, ensuring a consistent ground surface representation. These points are used to reconstruct the signed distance function (SDF) for the background, minimizing potential reconstruction errors. 

When dealing with dynamic foreground zones, our method assumes that only the ground surface \( G \) needs to be generated within these regions. However, ignoring dynamic objects entirely could lead to gaps in the ground's reconstruction. To address this, we first estimate an average height \( \bar{z}_G \) for the ground surface in the z-axis within a radius \( r \) around the dynamic mask \( D_i \). The average height is calculated as:

\begin{equation}
\bar{z}_G = \frac{1}{|P_G|} \sum_{\boldsymbol{p} \in P_G} z_p  
\end{equation}
where \( P_G \) is a set of ground points within the distance \( r \) around \( D_i \), and \( z_p \) is the z-coordinate of each point in \( P_G \).

To ensure a consistent ground surface representation, we then randomly generate LiDAR points \( \hat{P}_G = \{ \hat{\boldsymbol{p}}_1, \hat{\boldsymbol{p}}_2, \dots, \hat{\boldsymbol{p}}_n \} \) at the estimated height \( \bar{z}_G \). These points \( \hat{p}_j \) are positioned such that:

\begin{equation}
\hat{\boldsymbol{p}}_j = (x_j, y_j, \bar{z}_G), \quad j = 1, 2, \dots, n   
\end{equation}
where \( (x_j, y_j) \) are randomly sampled within \( D_i \).

These artificially generated points \( \hat{P}_G \) are then incorporated into the reconstruction process to generate the signed distance function (SDF) for the background. 

\textbf{Foreground SDF Loss Function:} 

The proposed foreground SDF Loss Function is designed to improve the accuracy and consistency of the Signed Distance Function (SDF) values in the NeRF-LOAM \cite{deng2023nerf} in highly dynamic environments where objects are moving.

For each LiDAR point $\boldsymbol{p}_i$, consider a dynamic region $D_i$ defined as a certain radius $r$ around the point. The SDF values within such dynamic regions are regularized to ensure that the geometric representation is smooth and accurate.

\begin{equation}
    \mathcal{L}_{\text{d}} = \frac{1}{|D_i|} \sum_{\boldsymbol{p}_j \in D_i} \left(\Psi(\boldsymbol{p}_j) - \frac{1}{|D_i|} \sum_{\boldsymbol{p}_k \in D_i} \Psi(\boldsymbol{p}_k)\right)^2
\end{equation}
Here, $\Psi(\boldsymbol{p}_j)$ is the SDF value at point $\boldsymbol{p}_j$ within the dynamic region $D_i$,   and the loss encourages the SDF values within the region to be consistent.

\subsection{Implicit neural scene representation}
\label{scene_representation}
% 多分辨率编码
\textbf{Multi-Resolution Encoding:}

% To address the inefficient memory usage caused by features in free space, Vox-fusion \cite{yang2022vox} uses an octree for dynamic voxel management, enabling fast voxel allocation and retrieval. NeRF-LOAM \cite{deng2023nerf} extends this approach to large-scale outdoor scenes and uses LiDAR-scanned point clouds as input for map reconstruction. NGLOD \cite{takikawa2021neural} achieves better reconstruction results by using multi-resolution feature fusion to train the network. 

Inspired by NGLOD \cite{takikawa2021neural}, we extend the octree structure in NeRF-LOAM \cite{deng2023nerf}. Specifically, we embed features only at the deepest $H$ levels of the octree to balance reconstruction quality and training speed. In other words, features are embedded only at levels $d \in \left \{ D_{max}-H+1, ..., D_{max} \right \}$, where we set $H = 3$, which was empirically found to provide a good balance, $D_{max}$ is the maximum depth of the octree. For a query point $\boldsymbol{p}$, its corresponding multi-resolution encoding is represented as:

\begin{equation} \label{eq:2}
    F_{\alpha }^{s}(\boldsymbol{p}) = \sum_{j=D_{max}-H+1}^{D_{max}} F_{\alpha }^{j}(\boldsymbol{p})   
\end{equation}

The embedding at each layer is obtained through trilinear interpolation of the embeddings at the eight vertices of the nodes in the current layer:

\begin{equation} \label{eq:2}
    F_{\alpha }^{j}(\boldsymbol{p}) = TriInpo(\boldsymbol{p}, \boldsymbol{e}_{1}^{j}, ..., \boldsymbol{e}_{8}^{j})
\end{equation}
where $TriInpo$ refers to trilinear interpolation, the parameters include the sample point $\boldsymbol{p}$ and the embeddings of the eight vertices $\boldsymbol{e}^{i}$ of the voxel containing the sample point. 
% The level index $v$ is defined as $v = D_{max} - 2 + j$.

\begin{figure*}[tp]
    \centering
    \vspace{8pt}
    \begin{subfigure}[b]{0.36\textwidth}
        \centering
        \includegraphics[width=\textwidth]{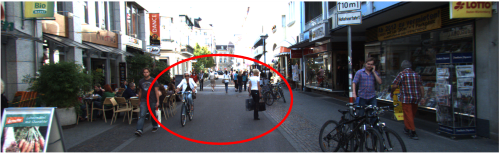}
        \caption{Dataset image in MOT19}
    \end{subfigure}
     \begin{subfigure}[b]{0.36\textwidth}
        \centering
        \includegraphics[width=\textwidth]{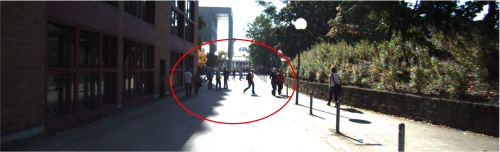}
        \caption{Dataset image in MOT26}
    \end{subfigure}

    \begin{subfigure}[b]{0.27\textwidth}
        \centering
        \includegraphics[width=\textwidth]{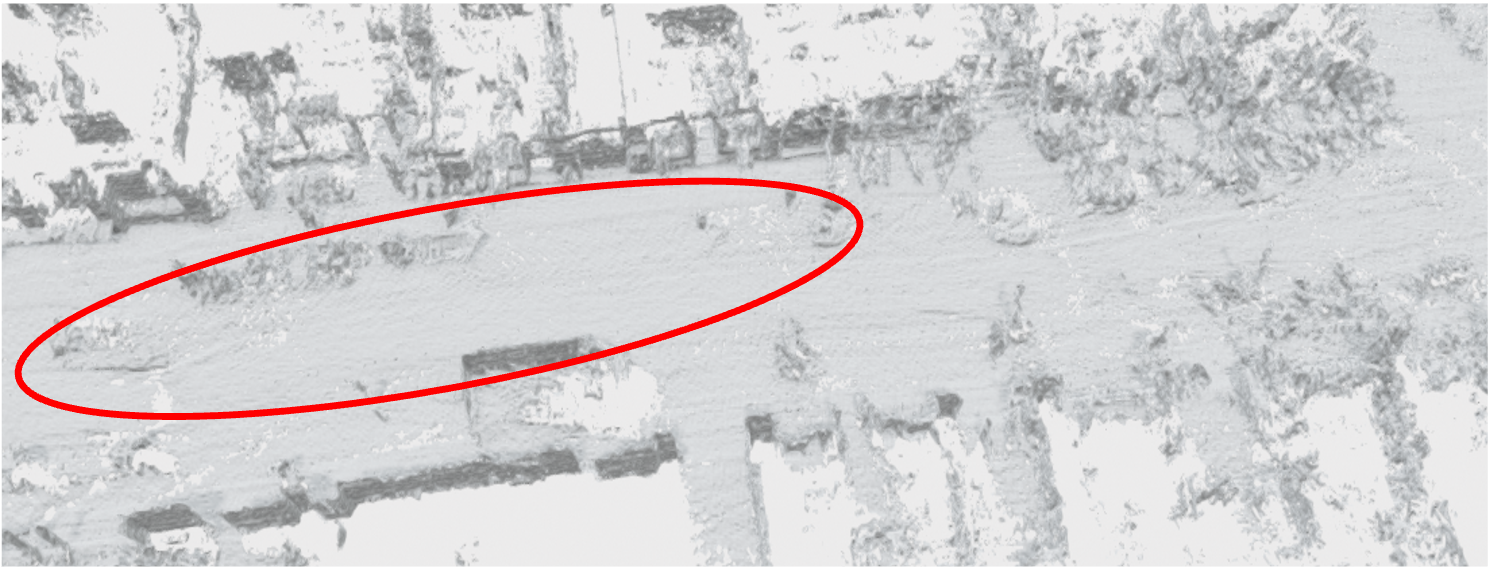}
        \caption{Proposed method in MOT19}
        \label{our19}
    \end{subfigure}
    \begin{subfigure}[b]{0.27\textwidth}
        \centering
        \includegraphics[width=\textwidth]{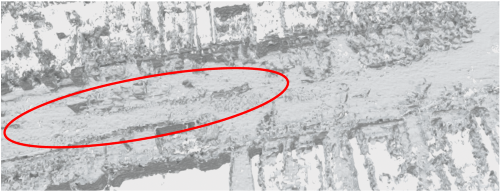}
        \caption{NeRF-LOAM in MOT19 }
        \label{nerf19}
    \end{subfigure}
     \begin{subfigure}[b]{0.27\textwidth}
        \centering
        \includegraphics[width=\textwidth]{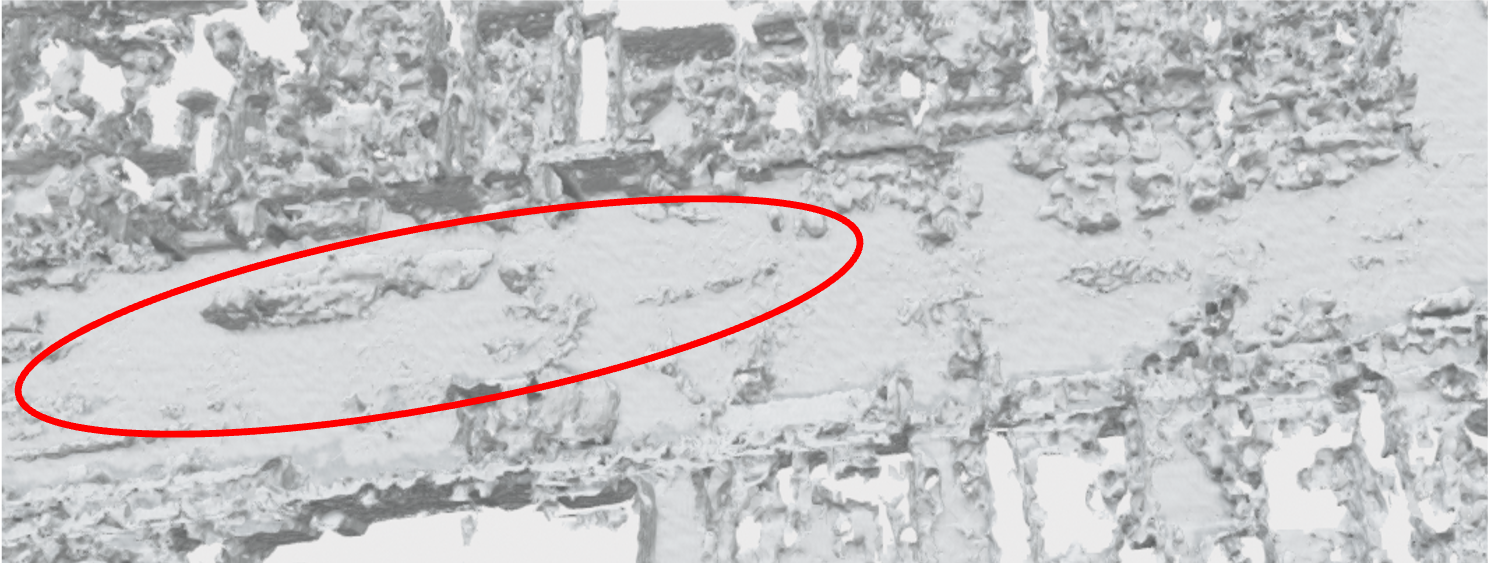}
        \caption{Pin-SLAM in MOT19 }
        \label{pin19}
        
    \end{subfigure}

     \begin{subfigure}[b]{0.27\textwidth}
        \centering
        \includegraphics[width=\textwidth]{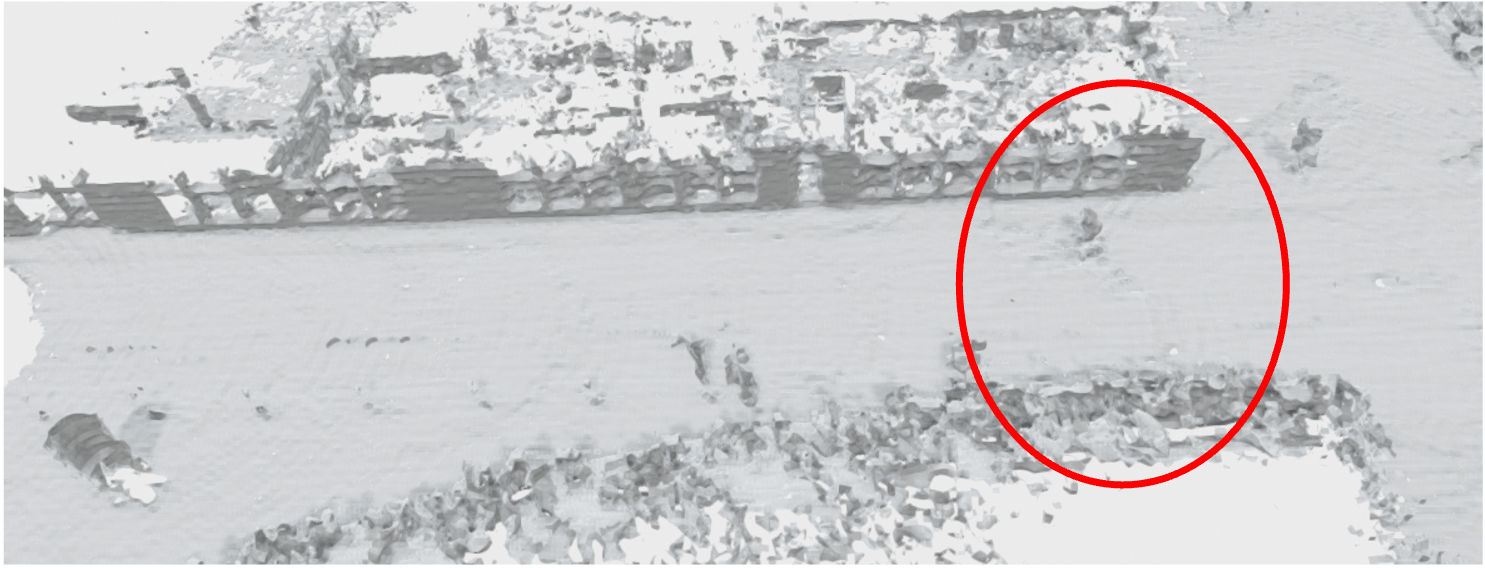}
        \caption{Proposed method in MOT26}
    \end{subfigure}
    \begin{subfigure}[b]{0.27\textwidth}
        \centering
        \includegraphics[width=\textwidth]{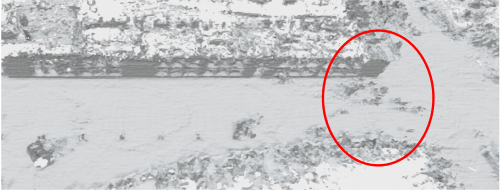}
        \caption{NeRF-LOAM in MOT26 }
    \end{subfigure}
    \begin{subfigure}[b]{0.27\textwidth}
        \centering
        \includegraphics[width=\textwidth]{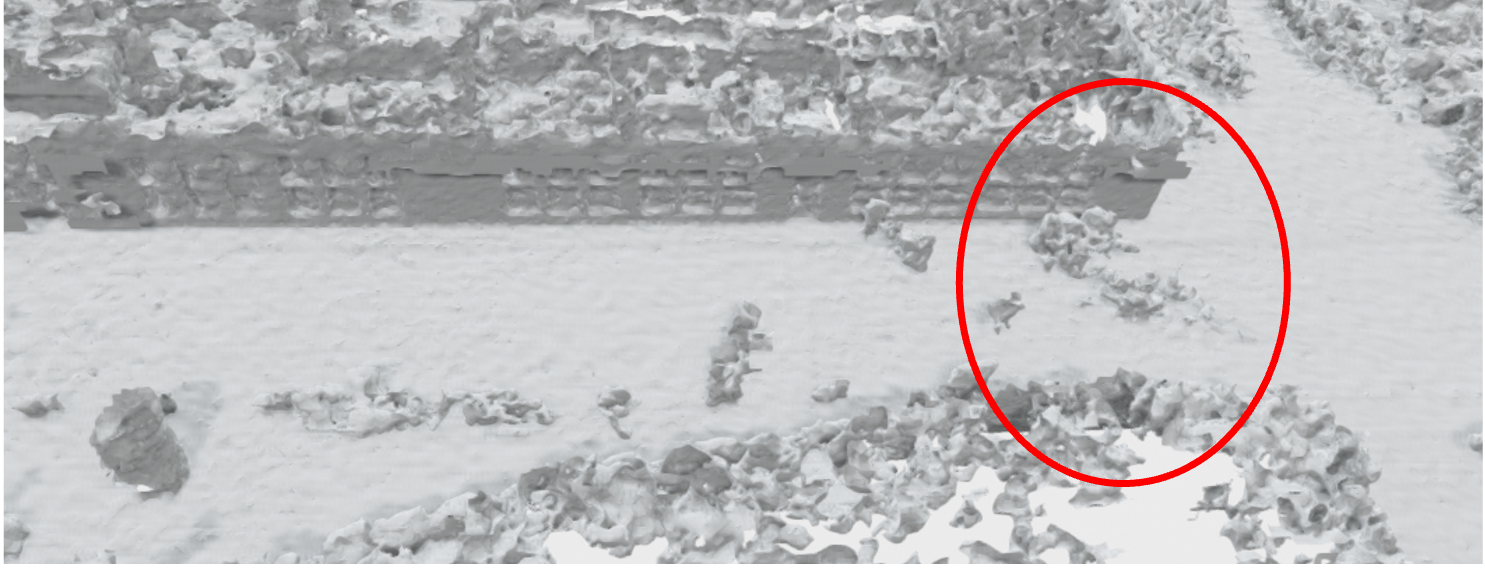}
        \caption{Pin-SLAM in MOT26 }
    \end{subfigure}
   
    \caption{3D reconstruction results among Proposed method, NeRF-LOAM \cite{deng2023nerf}, Pin-SLAM \cite{pan2024pin}. (a) and (b): Original images from the datasets MOT19 and MOT26. The red ovals highlight dynamic pedestrians of the scene. (c) to (h): These show how the proposed Method, NeRF-LOAM \cite{deng2023nerf}, and Pin-SLAM \cite{pan2024pin} perform on the MOT19 dataset. The grayscale images are reconstructions, and the red ovals indicate the areas of focus for comparison. }
    \label{our_exp}
\end{figure*}

\textbf{Fourier Features Positional Encoding:}

% NeRF-LOAM \cite{deng2023nerf} utilizes the embedding surrounding the sampled point to predict the SDF value at that point. While it has demonstrated considerable success in incremental mapping, challenges related to holes and smoothness remain.
% Despite the training, neural networks can gradually approximate the real values [1], the issues of holes in the reconstruction results and insufficient smoothness still persist.
% As demonstrated by Co-SLAM \cite{wang2023co}, in order to enhance the completion rate of the reconstruction, we integrate learnable features with positional encoding. 

Instead of using the frequency encoding adopted in NeRF \cite{mildenhall2021nerf} to encode the sampled points into a higher dimension, following the suggestion by Sun et al. \cite{sun20243qfp}, we input the coordinates of the sampled point, encoded and combined with the embedding of the point, into the neural SDF module to predict its SDF value. Given a query point $\boldsymbol{p}$, its corresponding Fourier positional encoding is represented: 
\begin{equation} \label{eq:1}
    \begin{split}
    \gamma(\boldsymbol{p}) = [sin(2\pi B_{1}\boldsymbol{p}), cos(2\pi B_{1}\boldsymbol{p}), ... ,\\
 sin(2\pi B_{k}\boldsymbol{p}), cos(2\pi B_{k}\boldsymbol{p})]^{\top } 
    \end{split}
\end{equation}
where $\mathbf{B_{i}}$ are coefficients $(i\in {1, 2, .. , k})$ sampled form an isotropic Gaussian distribution.
$\boldsymbol{B_{i} \sim }\mathcal{N} (0, \sigma ^{2} )$, and $\sigma$ is chosen for each task and dataset with a hyperparameter sweep. $k$ serves as a hyperparameter that determines the length of the positional encoding feature.

Finally, we concatenate the embedding after trilinear interpolation with the query point coordinates after Fourier positional encoding, and input it into the neural SDF module to predict the SDF value. The SDF value is represented as:
\begin{equation} \label{eq:2}
    \Psi(\boldsymbol{p}) = f( \gamma (\boldsymbol{p}), F_{\alpha }^{s}  ) 
\end{equation}where $\Psi$ is the predicted SDF value of a certain point.

\subsection{Optimization}
\label{loss}

The foreground loss can be integrated with the existing SDF loss $\mathcal{L}_s$, free space loss $\mathcal{L}_f$, and Eikonal loss $\mathcal{L}_e$ in NeRF-LOAM \cite{deng2023nerf}. The final loss function could be a weighted sum of these losses:

\begin{equation}
    \mathcal{L}_{\text{total}} = \lambda_s \mathcal{L}_s + \lambda_f \mathcal{L}_f + \lambda_e \mathcal{L}_e + \lambda_{\text{d}} L_{\text{d}}
\end{equation}where $\lambda_s$, $\lambda_f$, $\lambda_e$, and $\lambda_{\text{d}}$ are the weights that balance the contributions of each loss term.

\section{EXPERIMENTS}

In this section, we first elaborate on the detailed settings of the experiments (Section \ref{experiment_setup}). 
% We demonstrate in Section \ref{moving_obiect_removal} that our proposed method is more robust in dynamic scenes. 
As shown in Section \ref{map_results} and Section \ref{odometry_results}, the reconstruction results of our method are more competitive both qualitatively and quantitatively. Finally, in the ablation study, 
we demonstrated the effectiveness of each component in our proposed method. (Section \ref{ablation_study}).

\subsection{Experiment Setup}
\label{experiment_setup}

\subsubsection{\textbf{Baseline}}

Our method is compared with the current state-of-the-art methods NeRF-LOAM \cite{deng2023nerf} and Pin-SLAM\cite{pan2024pin} and Puma \cite{vizzo2021poisson} for the 3D reconstruction results and odometry. We also compare our method with the current state-of-the-art explicit reconstruction method VDBFusion \cite{vizzo2022vdbfusion} and the neural implicit reconstruction method SHINE-Mapping \cite{zhong2023shine}.

\begin{table*}[]
\caption{Quantitative evaluation of the reconstruction quality on the MaiCity and NewerCollege. The term 'w/o GT pose' refers to reconstructions using odometry-estimated poses, while 'w GT pose' refers to that using ground truth poses. 
% The best results are highlighted in bold, with the second-best results underlined. Arrows indicate whether lower ($\downarrow$) or higher ($\uparrow$) values are preferable.
}
\label{Table:quantitative_evaluation}
\centering
\resizebox{0.85\textwidth}{!}{
\begin{tabular}{ccccccc}
\hline
\hline
Dataset      &             & Method    & Comp.{[}cm{]}$\downarrow$   & Acc.{[}cm{]}$\downarrow$    & C-l1{[}cm{]}$\downarrow$    & F-score{[}\%{]}$\uparrow$ \\ \hline
             &             & Puma      & 9.14     & 7.89           & 8.51           & 68.04           \\
             &             & Pin-SLAM  & \underline{6.27}           & \underline{5.11}           & \underline{5.69}       & \underline{85.30} \\
             & w/o GT pose & NeRF-LOAM & 9.00          & 5.33     & 7.17     & 81.99     \\
             &             & Ours      & \textbf{5.71}  & \textbf{4.12}  & \textbf{4.91}  & \textbf{88.06}  \\ \cline{2-7}
MaiCity      &             & VDBFusion & 17.23          & \textbf{2.88}                 & 10.06           & 91.40           \\
             & w GT pose   & SHINE     & 4.24           & 3.88                 & 4.06           & 89.73           \\
             &             & NeRF-LOAM & \underline{4.11}           & 3.48       & \underline{3.79}           & \underline{93.30}     \\
             &             & Ours      & \textbf{3.68}              & \underline{3.41}           & \textbf{3.54}  & \textbf{94.11}  \\ \hline
             &             & Puma      & 71.91          & 15.30          & 43.60          & 57.27           \\
             &             & Pin-SLAM  & \underline{15.25}          & \underline{11.55}          & \underline{13.40}          & \textbf{82.08}   \\
             & w/o GT pose & NeRF-LOAM & 16.60          & 11.70          & 14.14          & 77.67     \\
             &             & Ours      & \textbf{15.05} & \textbf{10.91} & \textbf{12.98} & \underline{81.93}  \\ \cline{2-7}
NewerCollege &             & VDBFusion & 22.72          & \textbf{4.73}     & 13.72          & 91.13           \\
             & w GT pose   & SHINE     & {\underline{14.36}}    & 8.32           & 11.34          & 90.65           \\
             &             & NeRF-LOAM & 15.59          & \underline{6.86}  & {\underline{11.24}}    & {\underline{91.83}}     \\
             &             & Ours      & \textbf{11.14} & 7.63           & \textbf{9.26}  & \textbf{93.12}  \\ \hline\hline
\end{tabular}
}

\end{table*}

\subsubsection{\textbf{Datasets}}
We evaluate our method on three pubic Lidar datasets, including KITTI \cite{geiger2012we}, MaiCity \cite{vizzo2021poisson}, and Newer College \cite{ramezani2020newer} datasets. We use the MOT challenge in the KITTI dataset to evaluate our system with a highly dynamic outdoor dataset. MaiCity \cite{vizzo2021poisson} is a synthetic LiDAR dataset of outdoor urban environments. Newer College \cite{ramezani2020newer} is a real radar dataset on a university campus. 
KITTI \cite{geiger2012we} does not provide ground truth maps, therefore, we only provide the qualitative results of the reconstruction. The other two datasets provide registered dense point clouds as ground truth references for quantitative evaluation. 

\subsubsection{\textbf{Evaluation Metric}}

To ensure a fair comparison, for odometry, we use the Root Mean Square Error (RMSE) of the Absolute Trajectory Error (ATE) to evaluate. For mapping, we use the evaluation criteria employed in most methods \cite{deng2023nerf, vizzo2021poisson, zhong2023shine}, including accuracy, completion, Chamfer-L1 distance, and F-score. These metrics are calculated by comparing the ground truth and the predicted mesh.

% To ensure accurate mapping, we adopt the SHINE \cite{zhong2023shine} approach and evaluate the results using common reconstruction metrics, including accuracy, completion, Chamfer-L1 distance, and F-score. These metrics are calculated by comparing the ground truth and the predicted mesh.

\subsubsection{\textbf{Implementation Details}}
All experiments were run on an RTX 4090 GPU. The machine-learning-based moving object detection method employed was QD-3DT \cite{hu2022monocular}. The distance parameter \( r \) was set to 0.3 meters, meaning that points within a 0.3-meter radius around the moving object were considered reference ground points. For Fourier positional encoding, the variance parameter \( \sigma^2 \) was configured to 50, a value that proved effective across all datasets. The loss function parameter \( \lambda_L \) was set to 50. All other parameters were configured according to the settings specified in \cite{deng2023nerf}.

\begin{figure*}[tp]
    \centering
    \vspace{8pt}
    \begin{subfigure}[b]{0.2\textwidth}
        \centering
        \includegraphics[width=\textwidth]{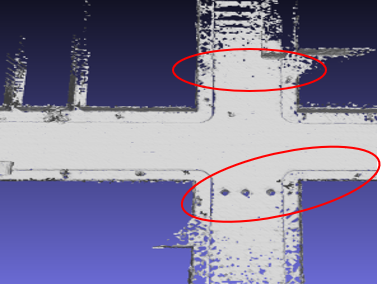}
        \caption{Ours in Mai}
        \label{mai01_our}
    \end{subfigure}
    \begin{subfigure}[b]{0.2\textwidth}
        \centering
        \includegraphics[width=\textwidth]{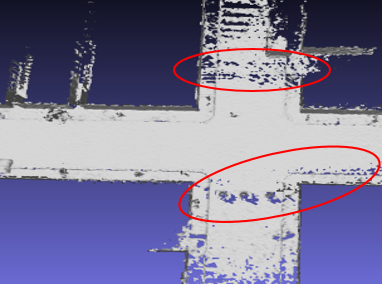}
        \caption{NeRF-LOAM in Mai}
        \label{mai01_ori}
    \end{subfigure}
    \begin{subfigure}[b]{0.2\textwidth}
        \centering
        \includegraphics[width=\textwidth]{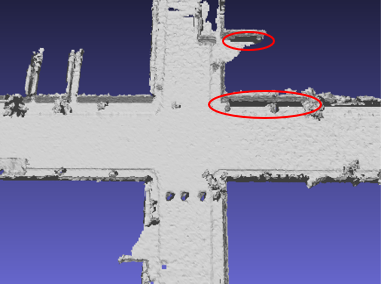}
        \caption{Pin-SLAM in Mai}
        \label{mai01_pin}
    \end{subfigure}
    \begin{subfigure}[b]{0.2\textwidth}
        \centering
        \includegraphics[width=\textwidth]{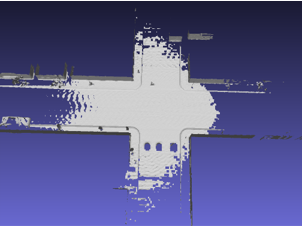}
        \caption{VDBFusion in Mai}
        \label{mai01_vdb}
    \end{subfigure}

    \begin{subfigure}[b]{0.2\textwidth}
        \centering
        \includegraphics[width=\textwidth]{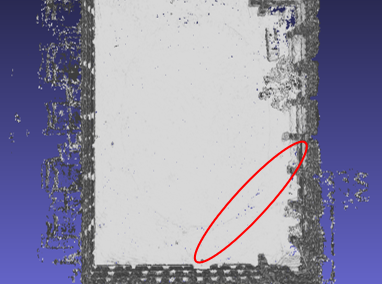}
        \caption{Ours in NCD}
        \label{ncd_our}
    \end{subfigure}
    \begin{subfigure}[b]{0.2\textwidth}
        \centering
        \includegraphics[width=\textwidth]{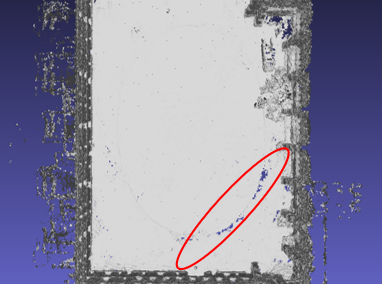}
        \caption{NeRF-LOAM in NCD}
        \label{ncd_ori}
    \end{subfigure}
    \begin{subfigure}[b]{0.2\textwidth}
        \centering
        \includegraphics[width=\textwidth]{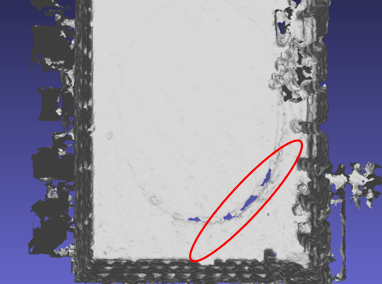}
        \caption{Pin-SLAM in NCD}
        \label{ncd_pin}
    \end{subfigure}
    \begin{subfigure}[b]{0.2\textwidth}
        \centering
        \includegraphics[width=\textwidth]{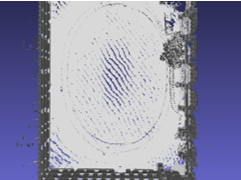}
        \caption{VDBFusion in NCD}
        \label{ncd_vdb}
    \end{subfigure}
    
    \caption{Qualitative visualization of the map quality on the MaiCity dataset (Mai) (top) and Newer College dataset (NCD) (down). The areas highlighted by the red ellipses emphasize the contributions of our proposed method. }
    \label{map_vis}
\end{figure*}

% \subsection{Moving Objects Removal}
% \label{moving_obiect_removal}

% The Fig.\ref{our_exp} highlights the differences between our proposed method, NeRF-LOAM \cite{deng2023nerf} and Pin-SLAM \cite{pan2024pin} in the context of dynamic object removal in outdoor street scenes. The proposed method excels in reconstructing scenes with moving objects, as evidenced by the clarity in the red-oval regions of the grayscale Fig.\ref{our_exp} (c) and (f). While NeRF-LOAM \cite{deng2023nerf} and Pin-SLAM \cite{pan2024pin} struggle to reconstruct the street that has moving objects, our proposed method maintains a more faithful representation of outdoor environment. 

\subsection{Map Results}
\label{map_results}
In this section, we first demonstrate the reconstruction results of our method compared to NeRF-LOAM \cite{deng2023nerf} and Pin-SLAM \cite{pan2024pin} on the MOT dataset. Subsequently, we present the quantitative performance of our method in comparison to Puma \cite{vizzo2021poisson}, Pin-SLAM \cite{pan2024pin} and NeRF-LOAM \cite{deng2023nerf} on the MaiCity \cite{vizzo2021poisson} and Newer College \cite{ramezani2020newer} datasets. None of these four methods rely on ground truth poses provided by the dataset as a reference. Finally, to provide a more comprehensive comparison of the reconstruction results, we compared our method with several reconstruction-focused methods, including VDBFusion \cite{vizzo2022vdbfusion} and SHINE \cite{zhong2023shine}. 

The Fig.\ref{our_exp} highlights the differences between our proposed method, NeRF-LOAM \cite{deng2023nerf} and Pin-SLAM \cite{pan2024pin} in the context of dynamic object removal in outdoor street scenes. The proposed method excels in reconstructing scenes with moving objects, as evidenced by the clarity in the red-oval regions of the grayscale Fig.\ref{our_exp} (c) and (f). While NeRF-LOAM \cite{deng2023nerf} and Pin-SLAM \cite{pan2024pin} struggle to reconstruct the street that has moving objects, our proposed method maintains a more faithful representation of outdoor environment. 

The quantitative evaluation results are shown in Table \ref{Table:quantitative_evaluation}. Although NeRF-LOAM \cite{deng2023nerf} achieves high reconstruction accuracy by separating ground points from non-ground points, its completeness is relatively low. In comparison, our method achieves higher completeness without compromising accuracy, attributed to the multi-resolution octree implementation and Fourier positional encoding, which allow the network to learn higher-dimensional information. 
% add
When using ground truth poses, VDBFusion \cite{vizzo2022vdbfusion} is reconstruction results are less complete compared to neural implicit methods due to its what-you-see-is-what-you-get storage characteristics. SHINE \cite{zhong2023shine} improves reconstruction quality through regularization and optimization strategies, but our hybrid encoding method still achieves more competitive results. 
Fig.\ref{map_vis} provides a qualitative analysis of the reconstructed maps, 
 Due to the explicit map representation, VDBFusion \cite{vizzo2022vdbfusion} is unable to reasonably complete unobserved areas, resulting in low completeness. Pin-SLAM \cite{pan2024pin} are neural points that can theoretically predict the SDF value at any location, but they are prone to generating unrealistic artifacts.
% 这里的内容整合到了前两段中
% To provide a more comprehensive comparison of the reconstruction results, we compare our method with several other methods focusing on reconstruction, and the evaluation results are show in Table \ref{Table:quantitative_evaluation}. The reconstruction method VDBFusion \cite{vizzo2022vdbfusion}, while achieving comparable or even superior performance to neural implicit reconstruction methods in terms of accuracy, falls far short in completeness compared to neural implicit reconstruction methods. Although the accuracy of our method is slightly lower in comparison, the completeness of the reconstruction has been significantly improved.

\subsection{Odometry Results}
\label{odometry_results}

\begin{table}[]
\caption{RMSE results of odometry. Mai for Maicity. NCD for NewerCollege. MOT for KITTI MOT chanlange dataset.}
\label{Table:Odometry}
\resizebox{0.48\textwidth}{!}{
\begin{tabular}{cccccc}
\hline \hline
Method    & Mai & NCD & MOT19 & MOT20 & MOT1-18\\ \hline
Pin-SLAM  &  \textbf{0.07}     &  \underline{0.09}   &  3.41 &  18.72 & 9.53\\
NeRF-LOAM &  0.19     &  0.14   &  \underline{3.32} & \underline{12.33} & \underline{9.36}\\
Ours      &  \underline{0.16}     &  \textbf{0.01}   &  \textbf{3.29} & \textbf{10.81} & \textbf{8.83}\\ \hline \hline
\end{tabular}
}
\end{table}

 Dynamic scenes pose significant challenges to odometry pose estimation. As shown in Table \ref{Table:Odometry}, our proposed method achieved competitive pose estimation accuracy especially in highly dynamic scenarios (MOT dataset) and NewerCollege.

\begin{table}[tp]
\caption{Ablation study of the proposed method. We present the quantitative results on the MaiCity01 sequence.}
\label{Table:ablation}
\resizebox{0.48\textwidth}{!}{
\begin{tabular}{cccc}
\hline
\hline
                & w/o FFE & w/o MF & Full  \\ \hline
Comp.{[}cm{]}$\downarrow$   & 9.59        & \underline{6.00}       & \textbf{5.71}  \\
Acc.{[}cm{]}$\downarrow$    & 4.82        & \underline{4.49}       & \textbf{4.12}  \\
C-l1{[}cm{]}$\downarrow$    & 7.21        & \underline{5.00}       & \textbf{4.91}  \\
F-score{[}\%{]}$\uparrow$ & 85.39        & \underline{85.72}       & \textbf{88.06} \\ \hline \hline
\end{tabular}
}
\end{table}

\subsection{Ablation Study}
\label{ablation_study}

We conducted ablation experiments to demonstrate the relative performance of our proposed method.

As shown in Fig.\ref{our_exp}, our dynamic object removal module effectively removes dynamic points and fills the gaps in the scene after removing the dynamic points. Moreover, Table \ref{Table:ablation} presents the quantitative evaluation of our method with different encoding strategies on Maicity \cite{vizzo2021poisson}. The completeness of the reconstruction results is poor without the Fourier feature positional encoding. The combination of multi-resolution octree structure and Fourier feature positional encoding achieves the overall best performance.


%%%%%%%%%%%%%%%%%%%%%%%%%%%%%%%%%%%%%%%%%%%%%%%%%%%%%%%%%%%%%%%%%%%%%%%%%%%%%%%%

%%%%%%%%%%%%%%%%%%%%%%%%%%%%%%%%%%%%%%%%%%%%%%%%%%%%%%%%%%%%%%%%%%%%%%%%%%%%%%%%

%%%%%%%%%%%%%%%%%%%%%%%%%%%%%%%%%%%%%%%%%%%%%%%%%%%%%%%%%%%%%%%%%%%%%%%%%%%%%%%%

\section{CONCLUSIONS}

In this paper, we present a 3D scene reconstruction system for dynamic outdoor environments, extending NeRF-LOAM \cite{deng2023nerf}. Our method excels in handling dynamic foregrounds by reconstructing only the ground surface while accurately modeling static backgrounds.

We use dynamic foreground masking and a novel ground height estimation method to ensure realistic reconstruction despite moving objects. The integration of a multi-resolution octree with Fourier feature positional encoding optimizes memory use and maintains scene quality.

% Experimental results on the MOT19 and MOT26 datasets show that our method surpasses NeRF-LOAM and Pin-SLAM in reconstruction accuracy, especially in dynamic environments. Figure \ref{our_exp} illustrates clear improvements in quality where dynamic objects are present. The ablation study highlights the effectiveness of Fourier encoding and multi-resolution octree structures in enhancing reconstruction accuracy and completeness. Additionally, our approach offers better pose estimation in dynamic settings, underscoring its robustness.


%%%%%%%%%%%%%%%%%%%%%%%%%%%%%%%%%%%%%%%%%%%%%%%%%%%%%%%%%%%%%%%%%%%%%%%%%%%%%%%%

%%%%%%%%%%%%%%%%%%%%%%%%%%%%%%%%%%%%%%%%%%%%%%%%%%%%%%%%%%%%%%%%%%%%%%%%%%%%%%%%

%%%%%%%%%%%%%%%%%%%%%%%%%%%%%%%%%%%%%%%%%%%%%%%%%%%%%%%%%%%%%%%%%%%%%%%%%%%%%%%%

\bibliographystyle{IEEEtran}
\bibliography{a}

\end{document}